\documentclass[letterpaper, 10 pt, conference]{ieeeconf}  

\IEEEoverridecommandlockouts                              

\usepackage{cite}
\usepackage{url}
\usepackage{algorithm}
\usepackage[noend]{algpseudocode}
\usepackage{subcaption}
\usepackage{mwe}
\usepackage{subcaption,siunitx,booktabs}
\usepackage{diagbox}
\usepackage{breqn}
\usepackage{hyperref}

\usepackage{amsthm,amsmath}
\usepackage{amssymb}
\let\labelindent\relax
\usepackage[shortlabels]{enumitem}
\theoremstyle{definition}

\usepackage{dsfont}
\usepackage{tikz-network}

\newtheorem{proposition}{Proposition}

\newtheorem{remark}{Remark}

\newtheorem{definition}{Definition}

\newcommand{\mc}[1]{\mathcal{#1}}

\newcommand{\mM}{\mathcal{M}}

\title{\LARGE \bf Evaluation Metrics for Object Detection for Autonomous Systems}

\author{Apurva Badithela$^{1,*}$, Tichakorn Wongpiromsarn$^{2}$ and Richard M. Murray$^{1}$ 
\thanks{We acknowledge funding from AFOSR Test and Evaluation Program, grant FA9550-19-1-0302, and from NSF grant CNS-2141153.}
\thanks{$^{1}$ Control and Dynamical Systems, California Institute of Technology.}
\thanks{$^{2}$ Department of Computer Science, Iowa State University.}
\thanks{*Corresponding author: A. Badithela \texttt{apurva@caltech.edu}}        
}

\begin{document}

\maketitle
\thispagestyle{empty}
\pagestyle{empty}

\begin{abstract}
This paper studies the evaluation of learning-based object detection models in conjunction with model-checking of formal specifications defined on an abstract model of an autonomous system and its environment. In particular, we define two metrics -- \emph{proposition-labeled} and \emph{class-labeled} confusion matrices -- for evaluating object detection, and we incorporate these metrics to compute the satisfaction probability of system-level safety requirements. While confusion matrices have been effective for comparative evaluation of classification and object detection models, our framework fills two key gaps. First, we relate the performance of object detection to formal requirements defined over downstream high-level planning tasks. In particular, we provide empirical results that show that the choice of a good object detection algorithm, with respect to formal requirements on the overall system, significantly depends on the downstream planning and control design. Secondly, unlike the traditional confusion matrix, our metrics account for variations in performance with respect to the distance between the ego and the object being detected. We demonstrate this framework on a car-pedestrian example by computing the satisfaction probabilities for safety requirements formalized in Linear Temporal Logic (LTL).
\end{abstract}


\section{Introduction}
\label{sec:intro}
\vspace{-1mm}
In safety-critical autonomous systems, such as self-driving cars, verifying learning-based modules with respect to formal requirements is necessary for safe operation. This presents the two challenges: i) correctly specifying the formal requirements to be verified, and ii) verification of the component. In this paper, we evaluate learning-based object detection and classification models with respect to system-level safety requirements encoded in Linear Temporal Logic (LTL). Inspired by the use of confusion matrices in machine learning\cite{koyejo2015consistent,wang2019consistent,yan2018binary,narasimhan2015consistent}, we define two new evaluation metrics for object detection and classification. We provide a framework for coupling these metrics with planning and control to provide a quantitative metric of safety at the complete system-level. 

In a typical autonomy software stack, the perception component is responsible for parsing sensor inputs to perform tasks such as object detection and classification, localization, and tracking, and the planning module is responsible for mission planning, behavior planning, and motion planning\cite{pendleton2017perception}. Further downstream in the autonomy stack, we have controllers to actuate the system to execute plans consistent with the planning module. Formal specification, synthesis, and verification of high-level plans for robotic tasks has been a successful paradigm in the last decade \cite{kress2009temporal,lahijanian2009probabilistic,wongpiromsarn2011tulip,kloetzer2008fully,raman2014model}. Some of this success can be attributed to the ease of specifying safety, progress, and fairness properties at the system-level\cite{baier2008principles, karlsson2020intention,karlsson2021encoding,hekmatnejad2019encoding,gassmann2019towards,shalev2017formal}. For instance, it is feasible to formally specify ``\textit{maintain safe longitudinal distance}" as a safety property for a self-driving car\cite{shalev2017formal,hekmatnejad2019encoding}. However, using the same formalism to evaluate perception can be challenging since it is difficult to formally characterize ``correct" performance (relative to human-level perception) for perception tasks such as object detection and classification\cite{dreossi2018semantic}. Consider classification of handwritten digits in the MNIST dataset\cite{deng2012mnist}, for which it is not possible to formally specify the pixel configuration that differentiates two digits. Likewise, it becomes impractical to express object detection and classification tasks in urban driving scenarios in temporal logic. 
\begin{figure}[h]
    \centering
    \includegraphics[width=\linewidth]{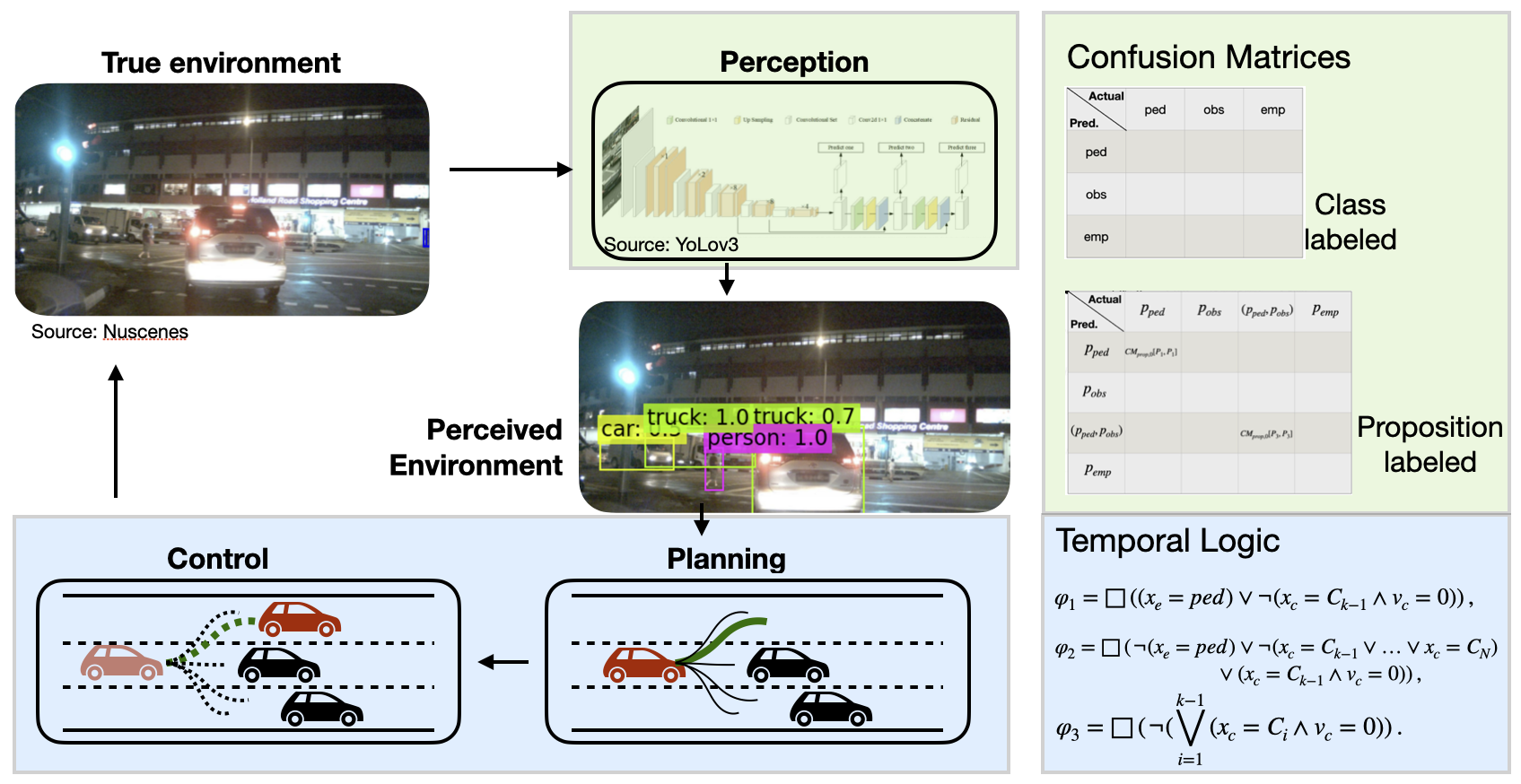}
    \caption{\small Planning and control is evaluated using temporal logic, and this analysis often excludes perception performance. Based on confusion matrices from computer vision, we propose new metrics for object detection, and outline a framework for accounting for perception performance in evaluating the overall system.}
    \label{fig:overview}
\end{figure}
In computer vision, object detection models are evaluated using metrics such as accuracy, precision and recall\cite{geron2019hands,koyejo2015consistent}. Several of these metrics are derived from the confusion matrix, which is constructed from model performance on an evaluation set\cite{geron2019hands,koyejo2015consistent}. Oftentimes, the training and evaluation of object detection models is subject to a qualitative assessment of whether it would lead to safe outcomes at the system-level. For example, researchers would expect that a model evaluated to have lower recall with respect to pedestrians would lead to better safety guarantees on the overall system. However, this was shown to not always be true --- lower recall resulting better performance with respect to some system-level safety requirements came at the cost of poorer performance in other safety requirements\cite{badithela2021leveraging}. 

In this work, first, we argue that the evaluation of object detection models should depend on the system-level specifications and the planner used downstream in the autonomy stack. For example, consider a cluster of pedestrians waiting at a crosswalk and an autonomous car with the requirement to come to a safe stop. If detecting one pedestrian vs. a group of pedestrians results in the same planning outcome (that is, the \emph{car safely stopping}), then accounting for every missed pedestrian would not accurately reflect the quantitative satisfaction of the safety requirement of the car. In particular, for quantitatively characterizing overall system satisfaction of safety, the choice of metrics for evaluating object detection must depend on the downstream planning logic. Secondly, object detection performance is better for objects closer to the autonomous system. Confusion matrices, as studied in computer vision, do not account for these two cases.

Our contributions in this work are the following. First, we study if distance does indeed result in a significant difference in the confusion matrix evaluation and eventual performance of the system with respect to safety requirements. For this, we use NuScenes\cite{caesar2020nuscenes} as a dataset to evaluate object detection models and construct the corresponding confusion matrices. We propose a distance-parametrized variation of the traditional confusion matrix to account for the effect of distance on object detection performance. Second, we define a proposition-labeled confusion matrix, and empirically show that is quantitatively more accurate with respect to the planner on discrete state examples. 


\section{Preliminaries}
\label{sec:prelim}
\vspace{-1mm}
In this section, we give an overview of temporal logic, which is useful in specifying system-level requirements formally. We then provide a background of performance metrics used to evaluate object detection and classification models in the computer vision community. Finally, we setup a discrete-state car-pedestrian system as a running example.
\vspace{-1mm}
\subsection{Temporal Logic for Specifying System-level Properties}
\label{sec:tl}
\vspace{-1mm}
\textbf{System Specification.}
\label{subsec:intro_example}
We use the term system to refer to refer to the autonomous agent and its environment. The agent is defined by variables \(V_A\), and the environment is defined by variables \(V_E\). The valuation of \(V_A\) is the set of states of the agent \(S_A\), and the valuation of \(V_E\) is the set of states of the environment \(S_E\). Thus, the states of the overall system is the set \(S := S_A\times S_E\). Let \(AP\) be a finite set of atomic propositions over the variables \(V_A\) and \(V_E\). An atomic proposition \(a\in AP\) is a statement that can be evaluated to \emph{true} or \emph{false} over states in \(S\).

We specify formal requirements on the system in LTL (see\cite{baier2008principles} for more details). An LTL formula is defined by (a) a set of atomic propositions, (b) logical operators such as: negation (\(\neg\)), conjunction (\(\wedge\)), disjunction \((\vee)\), and implication (\(\implies\)), and (c) temporal operators such as: next (\(\bigcirc\)), eventually (\(\diamond\)), always (\(\square\)), and until (\(\mathcal{U}\)). The syntax of LTL is defined inductively as follows: (a) An atomic proposition \(p\) is an LTL formula, and (b) if \(\varphi\) and \(\psi\) are LTL formulae, then \(\neg \varphi\), \(\varphi \vee \psi\), \(\bigcirc \varphi\), \(\varphi \,\mathcal{U}\, \psi\) are also LTL formulae. Further operators can be defined by a temporal or logical combination of formulas with the aforementioned operators. For an infinite trace \(\sigma = s_0s_1\ldots \), where \(s_i \in 2^{AP}\), and an LTL formula \(\varphi\) defined over \(AP\), we use \(\sigma \models \varphi\) to denote that \(\sigma\) satisfies \(\varphi\). 
For example, the formula \(\varphi = \square p\) represents that the atomic proposition \(p \in AP\) is satisfied at every state in the trace, i.e., \(\sigma \models \varphi\) if and only if \(p \in s_t, \forall t\). In this work, these traces \(\sigma\) are executions of the system, which we model using a Markov chain. 
\begin{definition}[Markov Chain\cite{baier2008principles}]
A discrete-time Markov chain is a tuple \(\mathcal{M} = (S, Pr, \iota_{init}, AP, L)\), where \(S\) is a non-empty, countable set of states, \(Pr:S\times S\rightarrow [0,1]\) is the \emph{transition probability function} such that for all states \(s \in S\), \(\Sigma_{s'\in S}Pr(s,s') = 1\), \(\iota_{init}:S \rightarrow [0,1]\) is the initial distribution such that \(\Sigma_{s\in S}\iota_{init}(s) = 1\), \(AP\) is a set of atomic propositions, and \(L:S\rightarrow 2^{AP}\) is a labeling function. The labeling function returns the set of atomic propositions that evaluate to true at a given state. Given an LTL formula \(\varphi\) (defined over \(AP\)) that specifies requirements of a system modeled by the Markov Chain \(\mathcal{M}\), the probability that a trace of the system starting from \(s_0 \in S\) will satisfy \(\varphi\) is denoted by \(\mathbb{P}_{\mathcal{M}}(s_0 \models \varphi)\). The definition of this probability function is detailed in\cite{baier2008principles}.
\end{definition}
\vspace{-1mm}
\subsection{Performance Metrics for Object Detection}
\vspace{-1mm}
\label{sec:pm}
We consider object detection to include both the detection and the classification tasks. In this section, we provide background on metrics used to evaluate performance with respect to these perception tasks.
Let the evaluation dataset \(\mathcal{D}=\{(f_i, b_i, d_i, x_i)\}_{i=1}^N\) consists of \(N\) objects in which each object the image frame token \(f_i\), the bounding box coordinates specified in \(b_i\), the distance of the object to ego \(d_i\) and the true annotation of the object \(x_i\). When a specific object detection algorithm is evaluated on \(\mc{D}\), each object has a predicted bounding box, \(\tilde{b}_i\), and predicted object class \(\tilde{x}_i\). We store these predictions in the set \(\mc{E}= \{(\tilde{b}_i, \tilde{x}_i)\}_{i=1}^N\). Now, we define the confusion matrix, a metric used to evaluate object detection models.
\begin{definition}[Confusion Matrix] Let \(\mc{D}\) be an evaluation set of objects along with their true and predicted labels. Let \(\mc{C}\) be a set of object classes in \(\mc{D}\), and let \(n\) denote the cardinality of \(\mc{C}\). The confusion matrix corresponding to the classes \(\mc{C}\) and  dataset \(\mc{D}\), and predictions \(\mc{E}\) is an \(n \times n\) matrix \(CM(\mc{C}, \mc{E}, \mc{D})\) with the following properties:
\begin{itemize}
    \item \(CM(\mc{C}, \mc{E}, \mc{D})[i,j]\) is the element in row $i$ and column $j$ of \(CM(\mc{C},\mc{E},\mc{D})\), and represents the number of objects that are predicted to have label \(c_i \in \mc{C}\), but have the true label \(c_j\in \mc{C}\), and 
    \item the sum of elements of the \(j^{th}\)-column of \(CM(\mc{C}, \mc{E}, \mc{D})\) is the total number of objects in \(\mc{D}\) belonging to class \(c_j \in \mc{C}\).
\end{itemize}
\label{def:cm}
\end{definition}
Several performance metrics for object detection and classification such as true positive rate, false positive rate, precision, accuracy, and recall can be identified from the confusion matrix\cite{koyejo2015consistent,wang2019consistent,geron2019hands}. 
\begin{remark}
In this work, we use \(c_n = empty\) (also abbreviated to \emph{emp} in figures) as an auxiliary class label in the construction of confusion matrices. If an object has a true label \(c_i\) but is not detected by the object detection algorithm, then this gets counted in \(CM(\mc{C}, \mc{E}, \mc{D})[n,i]\) as a false negative with respect to class \(c_i\). If the object was not labeled originally, but is detected and classified as a class \(c_i\), then it gets counted in \(CM(\mc{C}, \mc{E}, \mc{D})[i,n]\) as a false negative of the \emph{empty} class. We expect that in a properly annotated dataset, false negatives \(CM(\mc{C}, \mc{E}, \mc{D})[i,n]\) to be small. We ignore these extra detections in constructing the confusion matrix because by not being annotated, they are not relevant to the evaluation of object detection models.
\end{remark}
\vspace{-1mm}
\subsection{Example}
\begin{figure}
    \centering
    \includegraphics[scale=0.3]{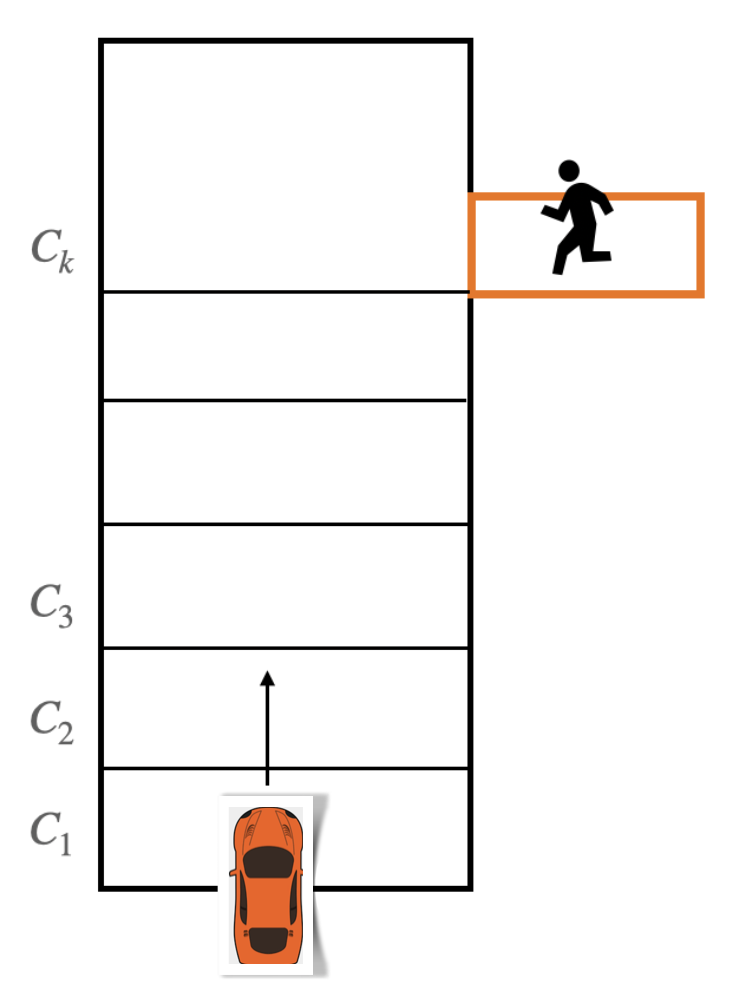}
    \caption{\small Running example of a car and pedestrian. If there is a pedestrian at crosswalk cell \(C_k\), that is, \(x_e \models ped\), then the car must stop at cell \(C_{k-1}\). Otherwise, it must not stop.}
    \label{fig:car_ped}
\end{figure}

\vspace{-1mm}
\label{sec:example}
Consider a car-pedestrian example, modeled using discrete transition system as illustrated in Figure~\ref{fig:car_ped}. The true state of the environment is denoted by \(x_e\), the state of the car comprises of its position and speed \((x_c, v_c)\). The safety requirement on the car is that it ``shall stop at the crosswalk if there is a waiting pedestrian, and not come to a stop, otherwise". 
The overall system specifications are formally expressed as safety specifications in equations~\ref{eq:spec1}-\ref{eq:spec3}. A more detailed description of this example can be found in \cite{badithela2021leveraging}. 
\begin{enumerate}[(S1)]
     \item If the true state of the environment does not have a pedestrian, i.e.
    \(x_e \neq ped\), then the car must not stop at \(C_{k-1}\).
     \item If \(x_e = ped\), the car must stop on \(C_{k-1}\).
     \item The agent should not stop at any cell \(C_i\), for all \(i \in \{1,\ldots, k-2\}\).\medskip
\end{enumerate}
\begin{equation}
\varphi_{1}= \square((x_e = ped)\vee \neg (x_c = C_{k-1} \wedge v_c = 0))\, , 
\label{eq:spec1}
\end{equation}
\begin{multline}
\varphi_{2}  = \square(\neg (x_e = ped)\vee \neg (x_c = C_{k-1}\vee \ldots \vee x_c = C_{N})\\ \vee (x_c = C_{k-1} \wedge v_c = 0))\, ,
\label{eq:spec2}
\end{multline}
\begin{equation}
    \varphi_3 = \square(\neg(\bigvee_{i = 1}^{k-1} (x_c = C_i \wedge v_c = 0)).
    \label{eq:spec3}
\end{equation} 
 The specifications (S\(1\)), (S\(2\)), and (S\(3\)) correspond to formulae $\varphi_1$, $\varphi_2$ and $\varphi_3$, respectively, and the specification the agent is expected to satisfy is \(\varphi:= \varphi_1 \land \varphi_2 \land \varphi_3\). 
\section{Metrics for Evaluating Object Detection}
\label{sec:metrics}
\vspace{-1mm}
In this section, we present two methods for constructing confusion matrices to evaluate object detection. The construction of the proposition-labeled and class-labeled confusion matrix is outlined in Algorithms~\ref{alg:cm_construct1} and \ref{alg:cm_construct2}, respectively.
While the confusion matrix provides useful metrics for evaluating object detection models, we would like to use these metrics in evaluating the performance of the end-to-end system with respect to formal constraints in temporal logic. In\cite{badithela2021leveraging}, we provided an algorithm that did system-level analysis by accounting for classification performance using the canonical confusion matrix. 
In this paper, we introduce two variants of the canonical confusion matrix as candidate metrics. For each confusion matrix, we evaluate the system end-to-end using the framework introduced in\cite{badithela2021leveraging}, and compare the results. 
\vspace{-1mm}
\subsection{Proposition-labeled Confusion Matrix}
\label{sec:prop_cm}
\vspace{-1mm}
In many instances, the planner does not need to correctly detect every object for high-level decision-making. For instance, for the planner to decide to stop for a group of pedestrians 20m away, the object detection does not need to correctly detect each and every pedestrian. In terms of quantifying system-level satisfaction of safety requirements, it is sufficient for the object detection to identify that there \emph{are} pedestrians 20m away, and not necessarily the precise number of pedestrians. Thus, we introduce the notion of using atomic propositions as class labels in the confusion matrix instead of the object classes themselves.
\begin{figure}
    \centering
    \includegraphics[scale=0.3]{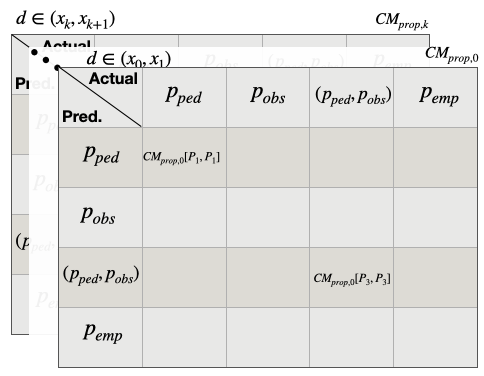}
    \caption{\small Proposition-labeled confusion matrix parametrized by distance to ego vehicle. The classes of interest are the atomic propositions: \(p_{ped} = \)``there is an object of class \emph{ped}",  obs \(=\) ``there is an object of class \emph{obs}", and \(p_{emp}=\)``there are no objects". The term (\(p_{ped}, p_{obs}\)) denotes the conjunction of atomic propositions \(p_{ped}\) and \(p_{obs}\).}
    \label{fig:prop_cm}
\end{figure}

Let \(n\) be the cardinality of the set of classes \(\mc{C}\) of objects of interest in the dataset \(\mc{D}\). Let \(p_i\) be the atomic proposition: ``there exists an object of class \(c_i\in \mc{C}\)", and, let \(\mc{P} = \{p_1,\ldots, p_n\}\) denote the set of all atomic propositions. For some \(k \in \{1,\ldots, k_{\text{max}}\}\), let \(0 < x_1 < \ldots < x_k < x_{k+1}< \ldots x_{k_{\text{max}}}\) denote progressively increasing distance from the autonomous vehicle. Let \(\mc{D}_k \subset \mc{D}\) be the subset of the dataset that includes objects that are in the distance range \(d_k = (x_k, x_{k+1})\) from the autonomous system. Let \(\mc{E}_k\) denote the predictions of the object detection algorithm corresponding to dataset \(\mc{D}_k\). For each parameter \(k\), we define the proposition-labeled confusion matrix \(CM_{prop,k}\): 
\begin{equation}CM_{prop,k} := CM(2^{\mc{P}}, \mc{E}_k, \mc{D}_k),\end{equation} 
with the classes characterized by the set \(2^{\mc{P}}\) and dataset \(\mc{D}_k\). Algorithm~\ref{alg:cm_construct1} shows the construction of the proposition-labeled confusion matrix.

\begin{algorithm}[H] 
\caption{Proposition-labeled Confusion Matrix}\label{alg1a} 
\begin{algorithmic}[1] \Procedure{$\mathtt{PropCM}$}{Dataset $\mc{D} = \{(f_i, b_i, d_i, x_i)\}_{i=1}^N$, Classes $\mc{C}$}
\State Define distance parameters: \(0 < d_1 < \ldots < d_{k_{\text{max}}}\) 
\State Run object detection algorithm to get predictions $\mc{E}$,
\State \(\mc{D}_0,\ldots, \mc{D}_{k_{\text{max}}} \leftarrow \mc{D}\) \Comment{Split dataset by distance}
\State \(\mc{E}_0,\ldots, \mc{E}_{k_{\text{max}}} \leftarrow \mc{E}\) \Comment{Split predictions}
\For  \(\quad c_j \in \mathcal{C}\)
\State \(p_j \leftarrow \text{object is of class } c_j\)
\EndFor
\State \(\mc{P} \leftarrow \bigcup_{j} p_j\) \Comment{Set of atomic propositions}
\For \(\quad k \in \{1,\ldots, k_{\text{max}}\}\)
\State Denote \(CM_{\text{prop}}(\mc{D}_k, \mc{E}_k, 2^{\mc{P}})\) as \(CM_{\text{prop},k}\)
\State \(CM_{\text{prop},k} \leftarrow \) zero matrix
\For \(\quad f_i \in \{f_1,\ldots, f_m\}\)  \Comment{Loop over images}
\State Group objects in \(\mc{D}_k\) with image token \(f_i\).
\State Group predictions in \(\mc{E}_k\) with image token \(f_i\).
\State \(P_i \leftarrow \) Predicted set of propositions
\State \(P_j \leftarrow \) True set of propositions
\State \(CM_{\text{prop},k}[P_i, P_j] \leftarrow CM_{\text{prop},k}[P_i, P_j] + 1\)
\EndFor
\EndFor
\State \(CM_{\text{prop}}(\mc{D}, \mc{E}, 2^{\mc{P}})\)\( = \{k: CM_{\text{prop}}(\mc{D}_k, \mc{E}_k, 2^{\mc{P}})\}\)
\State \textbf{return} \(CM_{\text{prop}}(\mc{D}, \mc{E}, 2^{\mc{P}})\)
\EndProcedure
\end{algorithmic}
\label{alg:cm_construct1}
\end{algorithm}
At every time step, for the distance range \((x_k, x_{k+1})\), the true environment is associated by a set of atomic propositions evaluating to true. Suppose, there is a pedestrian and a trash can in the distance range \((x_k, x_{k+1})\), then the true class of the environment in the confusion matrix corresponds to \(x_e=(p_{\text{ped}}, p_{\text{obs}})\). Note that for every possible environment, there is only one corresponding class in the confusion matrix. Thus, for a given true environment, the predicted class of the environment at distance \(d_k\) could be any one of the elements in \(2^{\mc{P}}\). Therefore, the outcomes of a environment at each time step are \(Outc = 2^{\mc{P}}\).
The tuple \((Outc, 2^{Outc})\) forms a \(\sigma\)-algebra for defining a probability function over the confusion matrix. Since the set \(Outc\) is countable, we can define a probability function \(\mu: Outc \rightarrow [0,1]\) such that \(\sum_{e \in Outc} \mu(e) = 1\). For a confusion matrix \(CM_{prop,k}\) with classes in the set \(Outc\), and for every true environment class \(P_j\), we can define the corresponding probability function \(\mu_k(\cdot, P_j): Outc \rightarrow 2^{Outc}\) as follows,
\begin{equation}
    \forall P_i \in 2^{\mc{P}}, \quad \mu_k(P_i, P_j) = \frac{CM_{prop,k}[P_i, P_j]}{\sum_{l=1}^{|2^{\mc{P}}|} CM_{prop,k}[P_l, P_j]},
\end{equation}
where $CM_{prop,k}[P_i, P_j]$ is the element of the confusion matrix corresponding to the row labeled by set of atomic propositions \(P_i\) and column labeled by set of atomic propositions labeled \(P_j\).
That is, we can define \(2^{|\mc{P}|}\) different probability functions, one for each possible true environment, for the confusion matrix \(CM_{prop,k}[P_i, P_j]\). Thus, for each distance parameter \(k\), we can define a probability function \(\mu_k\) that characterizes the probability of detecting one set of propositions \(P_i\), given that the true environment corresponds to \(P_j\) propositions. This helps us formally define the state transition probability of the overall system as following. 
\begin{definition}[Transition probability function for proposition-labeled confusion matrices]
Let \(x_e\) be the true state of the environment corresponding to the proposition \(P_j\), and let \(s_{a,1}, s_{a,2} \in S\) be states of the car. Let \(O(s_1, s_2)\) denote the set of all observations of the environment that prompt the system to transition from \(s_1 = (s_{a,1}, x_e)\) to \(s_2 = (s_{a,2}, x_e)\). At state \(s_1\), let \(k\) be the distance parameter of objects in the environment causing the agent to transition from state \(s_{a,1}\) to state \(s_{a,2}\). The corresponding confusion matrix is \(CM_{prop,k}\). Then, the transition probability from state \(s_1\) to \(s_2\) is defined as follows,
\begin{equation}
    Pr(s_1, s_2) := \sum_{P_i \in O(s_1, s_2)} \mu_k (P_i, P_j).
\end{equation}
\end{definition}
\begin{remark}
For brevity, we choose to assume that object at a specific distance range dominate the system transition from \(s_1\) to \(s_2\). However, this choice depends on the agent's planner and can be extended to the case in which objects at multiple distances are dominating influences. For that, we would consider the probability of detecting correct propositions in one distance range independent of correctly detecting propositions for a different distance range.
\end{remark}
\vspace{-1mm}
\subsection{Class-labeled Confusion Matrix parametrized by distance}
\label{sec:class_cm}
\vspace{-1mm}
This performance metric builds on the class-labeled confusion matrix defined in Definition~\ref{def:cm}. As denoted previously, let \(\mc{C}=\{c_1, \ldots, c_n\}\) be the set of different classes of objects in dataset \(\mc{D}\). For every object in \(\mc{D}_k\), the predicted class of the object will be one of the class labels \(c_1, \ldots, c_n\). For each distance parameter \(k\), we define the class-labeled confusion matrix as follows:
\begin{equation}
    CM_{class,k}:= CM(\mc{C},\mc{E}_k, \mc{D}_k).
\end{equation}
Algorithm~\ref{alg:cm_construct2} shows the construction of the class-labeled confusion matrix.
Therefore, the outcomes of the object detection algorithm for any given object will be defined by the set \(Outc = \{c_1,\ldots, c_n\}^{m}\), where \(m\) is the total number of objects in the true environment in distance \(d_k\). The tuple \((Outc, 2^{Outc})\) forms a \(\sigma\)-algebra for defining a probability function over the confusion matrix \(CM_{class,k}\). Similar to the definition of a probability function in the prior section, for every true environment \(c_j\), the probability function \(\mu_k(\cdot, c_j): Outc \rightarrow [0,1]\) can be defined as follows,
\begin{equation}
    \mu_k(c_i, c_j) := \frac{CM_{class,k}[c_i, c_j]}{\sum_{l=1}^n CM_{class,k}[c_l, c_j]}.
\end{equation}

\begin{definition}[Transition probability function for class-labeled confusion matrix]
Let the true environment be represented as a tuple \(x_e\) corresponding to class labels in the region \(d_k\) (class labels can be repeated in a tuple \(x_e\) when multiple objects of the same class are in region \(d_k\)). Let \(x_e\) be a tuple comprising of true lables of objects in the environment, and let \(s_{a,1}, s_{a,2} \in S\) be states of the car. Let \(O(s_1, s_2)\) denote the set of all observations of the environment that prompt the system to transition from \(s_1 = (s_{a,1}, x_e)\) to \(s_2 = (s_{a,2}, x_e)\). Likewise, let the tuple \(y_e\) represent the observations of the environment. Then, the transition probability function from state \(s_1\) to \(s_2\) is defined as follows,
\begin{equation}
    Pr(s_1, s_2) := \sum_{y_e \in O(s_1, s_2)} \prod_{i=1}^{|y_e|} \mu_k (y_e(i), x_e(i)).
    \label{eq:indep_prod}
\end{equation}
\end{definition}
For both transition probability functions, we can check (by construction) that \(\forall s_1 \in S, \sum_{s_2} Pr(s_1, s_2) = 1\).
In the running example, if the sidewalk were to have another pedestrian and a non-pedestrian obstacle, then the probability of detecting each object is considered independently of the others. This results in the product of probabilities \(\mu_k (\cdot, x_e(i))\) in equation~\eqref{eq:indep_prod}. 

\begin{algorithm}[H] 
\caption{Class-labeled Confusion Matrix}\label{alg1b} 
\begin{algorithmic}[1] \Procedure{$\mathtt{ClassCM}$}{Dataset $\mc{D} = \{(f_i, b_i, d_i, x_i)\}_{i=1}^N$, Classes $\mc{C}$}
\State Run object detection algorithm to get predictions $\mc{E}$,
\State \(\mc{D}_0,\ldots, \mc{D}_{k_{\text{max}}} \leftarrow \mc{D}\) \Comment{Split dataset by distance}
\State \(\mc{E}_0,\ldots, \mc{E}_{k_{\text{max}}} \leftarrow \mc{E}\) \Comment{Split predictions}
\For \(\quad k \in \{1,\ldots, k_{\text{max}}\}\)
\State Denote \(CM_{\text{class}}(\mc{D}_k, \mc{E}_k, 2^{\mc{P}})\) as \(CM_{\text{class},k}\)
\State \(CM_{\text{class},k} \leftarrow \) zero matrix
\For \(\quad f_i \in \{f_1,\ldots, f_m\}\)  \Comment{Loop over images}
\For \(\quad\)  object in \(\mc{D}_k\) 
\State \(c_i \leftarrow \) Predicted class label of object
\State \(c_j \leftarrow \) True class label of object in \(\mc{E}_k\) 
\State \(CM_{\text{class},k}(c_i, c_j) \leftarrow CM_{\text{class},k}(c_i, c_j) + 1\)
\EndFor
\EndFor
\EndFor
\State \(CM_{\text{class}}(\mc{D}, \mc{E}, 2^{\mc{P}})\)\( = \{k: CM_{\text{class}}(\mc{D}_k, \mc{E}_k, 2^{\mc{P}})\}\)
\State \textbf{return} \(CM_{\text{class}}(\mc{D}, \mc{E}, 2^{\mc{P}})\)
\EndProcedure
\end{algorithmic}
\label{alg:cm_construct2}
\end{algorithm}
\vspace{-1mm}
\subsection{Markov Chain Construction\cite{badithela2021leveraging}}
\vspace{-1mm}
For each confusion matrix, we can synthesize a corresponding Markov chain of the system state evolution as per Algorithm~\ref{alg:mc_construct1}. As a result of prior work shown in\cite{badithela2021leveraging}, using off-the-shelf probabilistic model checkers such as Storm\cite{dehnert2017storm}, we can compute the probability that the trace of a system satisfies its requirement, \(\mathbb{P}(s_0 \models \varphi)\), by evaluating the probability of satisfaction of the requirement \(\varphi\) on the Markov chain. Let \(O(x_e)\) be the set of all possible predictions of true environment state \(x_e\) by the the object detection model. Given a system controller \(K: S \times O(x_e) \rightarrow S\) takes as inputs the current state of the agent and the environment, \(s_0 \in S\), and environment state predictions \(y_e \in O(x_e)\) passed down by the object detection model. Based on the predictions, it chooses an end state \(s_f \in S\) by actuating the agent. For the car pedestrian running example, we use a correct-by-construction controller for the specifications~\eqref{eq:spec1}-\eqref{eq:spec3} corresponding to each true environment \(x_e\in\{ped, obs, emp\}\). At each time step, the agent makes a new observation of the environment (\(y_e\)) and chooses an action with the controller corresponding to \(y_e\). For the running example, the runtime for constructing the Markov chain was on the order of \(0.1\) seconds.
\begin{algorithm}[H] 
\caption{Markov Chain Construction}\label{alg2a} 
\begin{algorithmic}[1] \Procedure{$\mathtt{Markov\text{ } Chain}$}{$S,K,O(x_e),CM,x_e$}
\State \(Pr(s, s') = 0,\, \forall s,s'\in S\) 
\For{$s_o \in S$}
\State $\iota_{init}(s_0) = 1$ 
\For{$y_e \in O(x_e)$} 
      \State $s_f\leftarrow K(s_o, y_e)$  \Comment{Controller}
      \State Identify \(k\) relevant to \(s_0\) to \(s_f\) transition 
      \State $p\leftarrow CM_k[y_e,x_e]$ \Comment{Distance parametrized confusion matrix}
      \State $Pr(s_o, s_f) \leftarrow Pr(s_o, s_f) + p$
   \EndFor\label{obs_loop}
\EndFor\label{state_loop}
   \State \textbf{return} $\mM = (S, Pr, \iota_{init}, AP, L)$
\EndProcedure
\end{algorithmic}
\label{alg:mc_construct1}
\end{algorithm}

\begin{proposition}
Suppose we are given: i) \(\varphi\) as a temporal logic formula over system states \(S\), ii) true state of the environment \(x_e\), iii) agent initial state \(s_{a,0}\), and iv) a Markov chain \(\mathcal{M}\) constructed via Algorithm~\ref{alg:mc_construct1} for the distance parameterized confusion matrices constructed from Algorithm~\ref{alg:cm_construct1} or Algorithm~\ref{alg:cm_construct2}, then \(\mathbb{P}(s_0\models \varphi)\) is equivalent to computing \(\mathbb{P}_{\mathcal{M}}(s_0 \models \varphi)\), where \(s_0 = (s_{a,0}, x_e)\). 
\label{prop:p1}
\end{proposition}

\section{Simulation Results}
\vspace{-1mm}
\label{sec:results}
We present the distance-parametrized proposition-labeled and class-labeled confusion matrices for a pre-trained YoLo-v3 model\cite{redmon2018yolov3} evaluated on a portion of the NuScenes dataset\cite{caesar2020nuscenes}. Furthermore, we compute the probability of satisfaction that the car will meet its safety requirements in the car-pedestrian example.
\begin{figure}
\centering
   \includegraphics[width=0.8\linewidth]{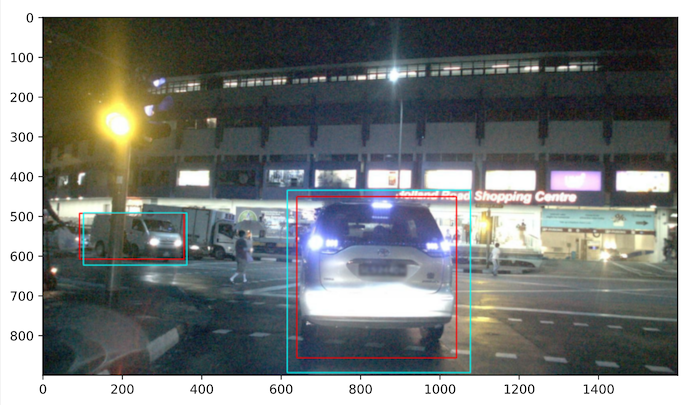}
   \caption{\small YoLo-v3 detecting the car and truck as objects of class \(obs\), but not detecting the pedestrians. The red boxes denote the YoLo predictions and the light blue boxes denote the corresponding ground truth annotations for correctly detected objects.}
   \label{fig:nusc_obj} 
\end{figure}
The pre-trained YoLov3\cite{redmon2018yolov3} model was trained on the MS COCO dataset\cite{lin2014microsoft}. Note that we chose to evaluate a model on an evaluation set different from the source of the training set.
\begin{figure}
\centering
   \includegraphics[width=0.8\linewidth]{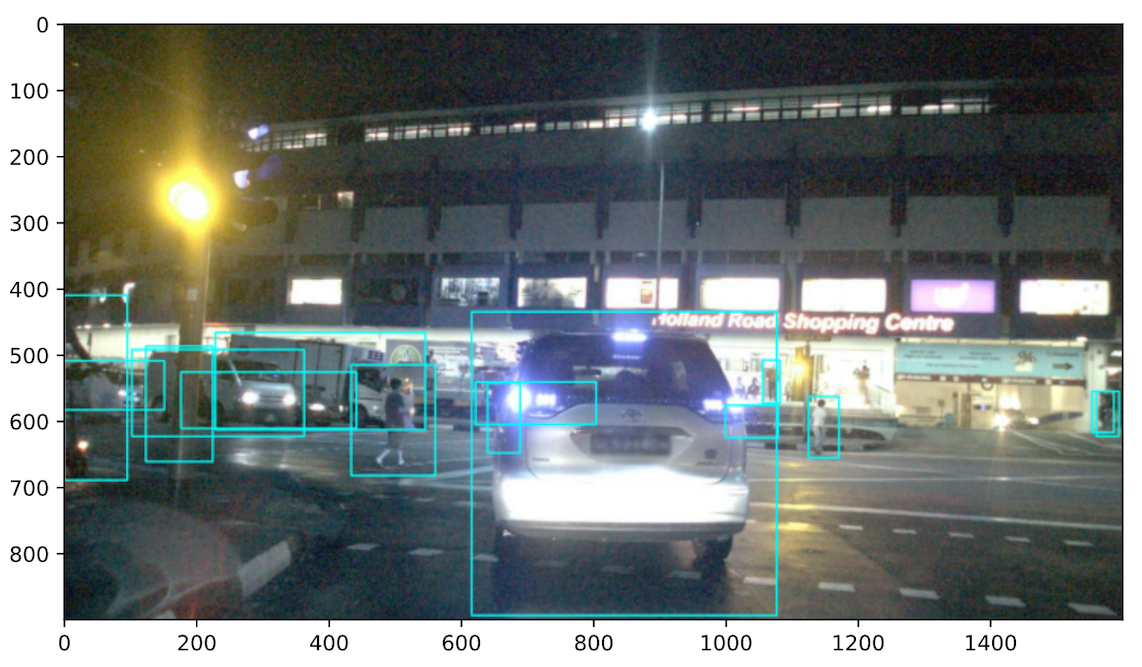}
   \caption{\small Ground truth annotations of a NuScenes frame with 3D annotation boxes projected to 2D.}
   \label{fig:nusc_gt}
\end{figure}
 We used the first 85 scenes of the Full NuScenes v1.0 dataset. Each scene is 20 seconds long, with 3D object annotations made at 2 Hz for 23 different classes. All objects with Nuscenes annotation ``human" are clustered under the class \(ped\), and all objects annotated as ``vehicle", ``static obstacle", and moving obstacle are annotated as ``obs". We use all \(40\) frames from each scene in our dataset \(\mc{D}\). We also project the 3D bounding boxes to 2D to match the predicted object detections from the YoLov3 model.

The class-labeled confusion matrix for objects less than 10 m from the autonomous vehicle is given in Figure~\ref{tab:cb} and the proposition-labeled confusion matrix for objects less than 10 m from the autonomous vehicle is given in Figure~\ref{tab:pb}. Due to space restrictions, we have the rest of the distance parametrized confusion matrices on this GitHub repository\footnote{\href{https://github.com/abadithela/Dist-ConfusionMtrx}{https://github.com/abadithela/Dist-ConfusionMtrx}}

\begin{figure}
\centering
\begin{tabular}{|l|rrr|}
\hline
\diagbox{\textit{Pred}}{\textit{True}}   &   \textit{ped} &   \textit{obs} &  \textit{emp} \\
\hline
 \textit{ped} &                31 &               0 &            0 \\
 \textit{obs}   &                 0 &             191 &            0 \\
\textit{emp}      &               127 &             734 &         3227 \\
\hline
\end{tabular}
\caption{\small Class-labeled Confusion matrix for distance \(d \leq  \) 10.0\\}
\label{tab:cb}
\end{figure}


\begin{figure}
\centering
\begin{tabular}{|l|rrrr|}
\hline
\diagbox{\textit{Pred}}{\textit{True}} &   \textit{ped} &   \textit{obs} &   \textit{ped}, \textit{obs} &  \textit{emp} \\
\hline
 \textit{ped}            &                22 &               0 &                            5 &            0 \\
 \textit{obs}              &                 0 &             184 &                            4 &            0 \\
 \textit{ped}, \textit{obs} &                 0 &               0 &                            0 &            0 \\
\textit{emp}                 &                63 &             354 &                           13 &         3227 \\
\hline
\end{tabular}
\caption{\small Proposition-labeled Confusion matrix for distance \(d \leq  \) 10.0. As expected, the sum total of samples for each label are no larger than for the corresponding label in the class-based confusion matrix.}
\label{tab:pb}
\end{figure}

\begin{figure}
\begin{subfigure}{0.45\columnwidth}
       \centering
       \includegraphics[width=\linewidth]{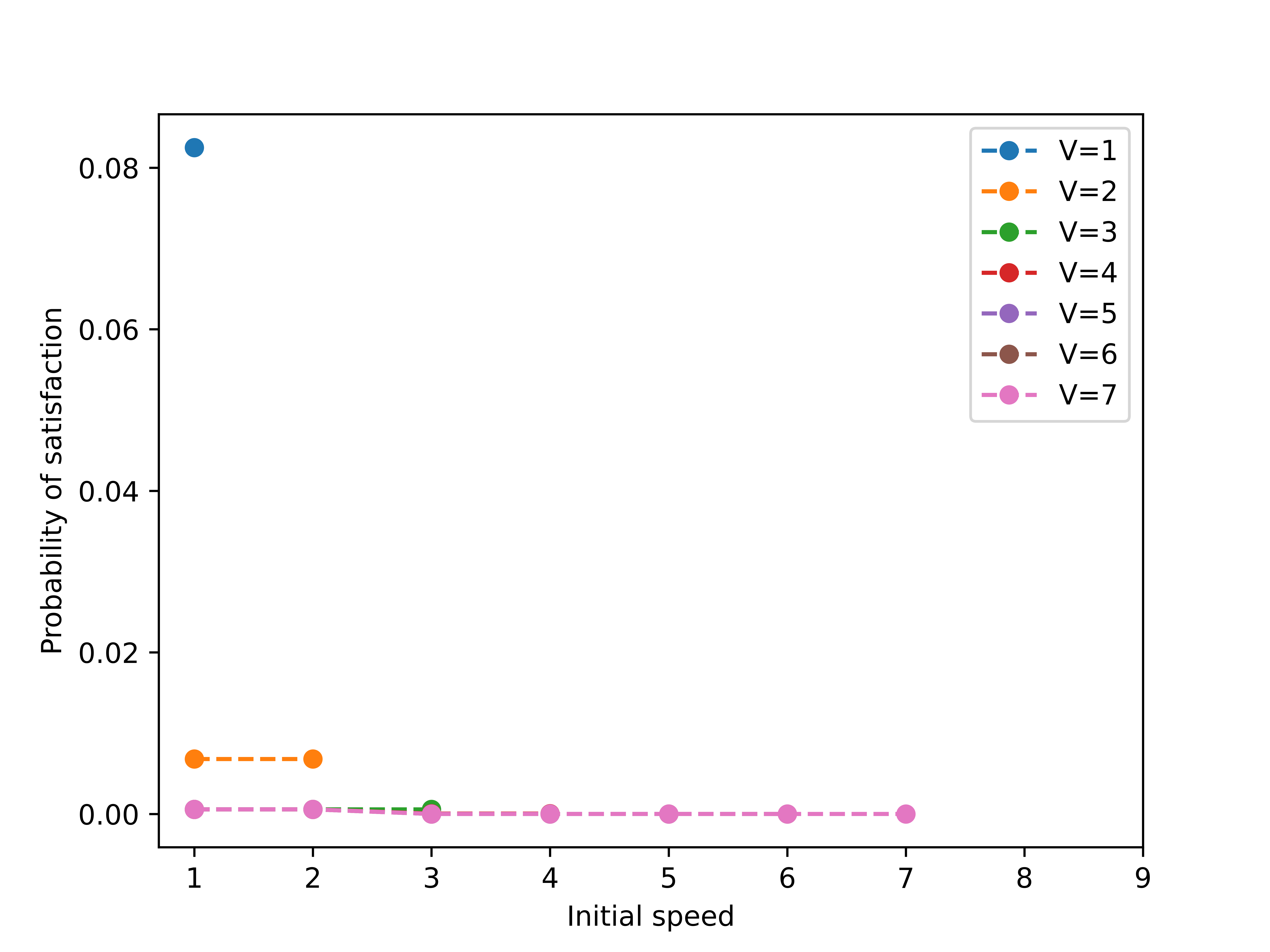}
       \caption{\footnotesize Class-labeled confusion matrix}
       \label{fig:class_plot}
   \end{subfigure} 
    \hfill
\begin{subfigure}{0.45\columnwidth}
       \centering
       \includegraphics[width=\linewidth]{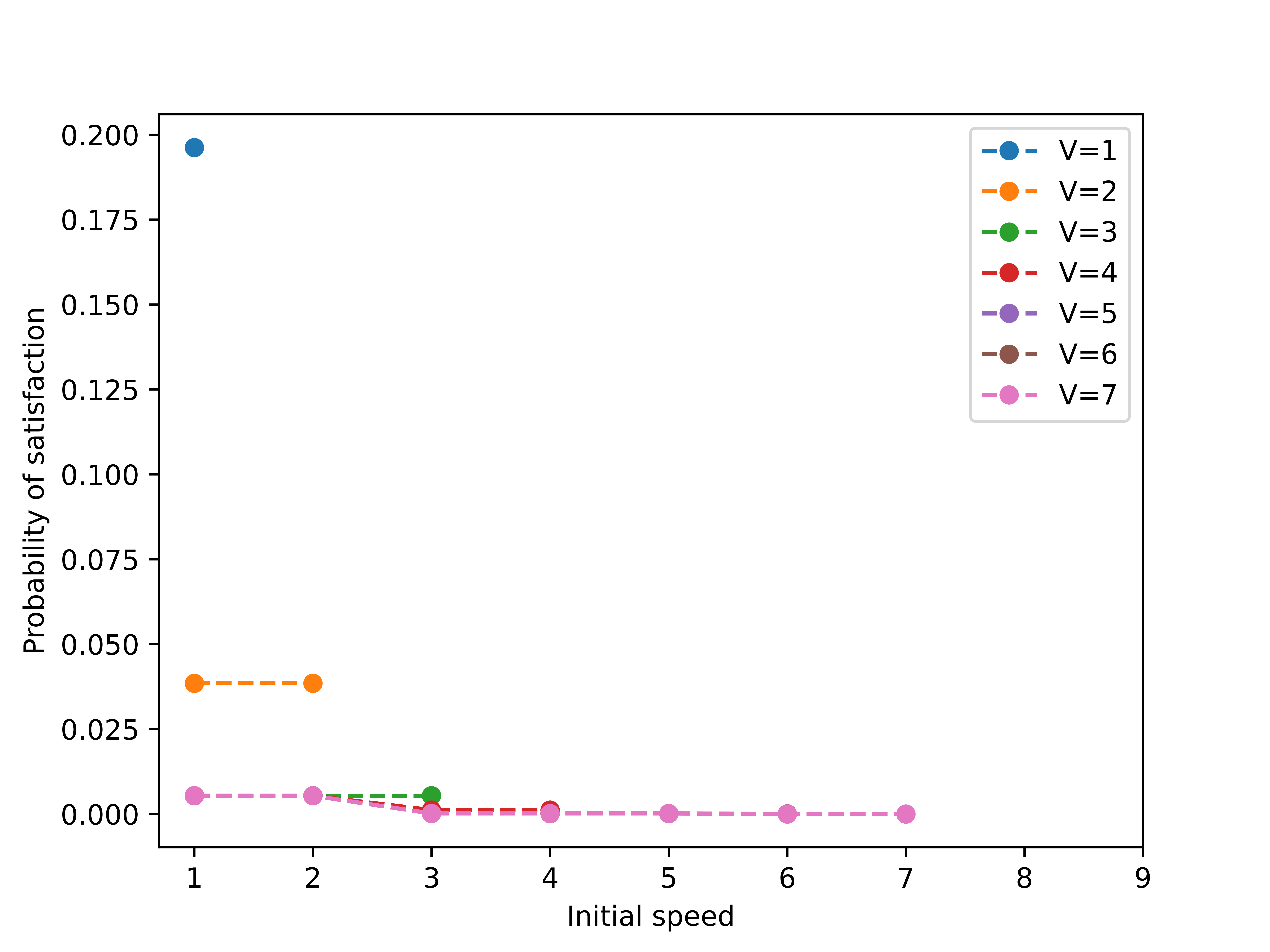}
       \caption{\footnotesize Class-labeled confusion matrix parametrized by distance}
       \label{fig:dist_class_plot}
   \end{subfigure} 
 \\
\begin{subfigure}{0.45\columnwidth}
       \centering
       \includegraphics[width=\linewidth]{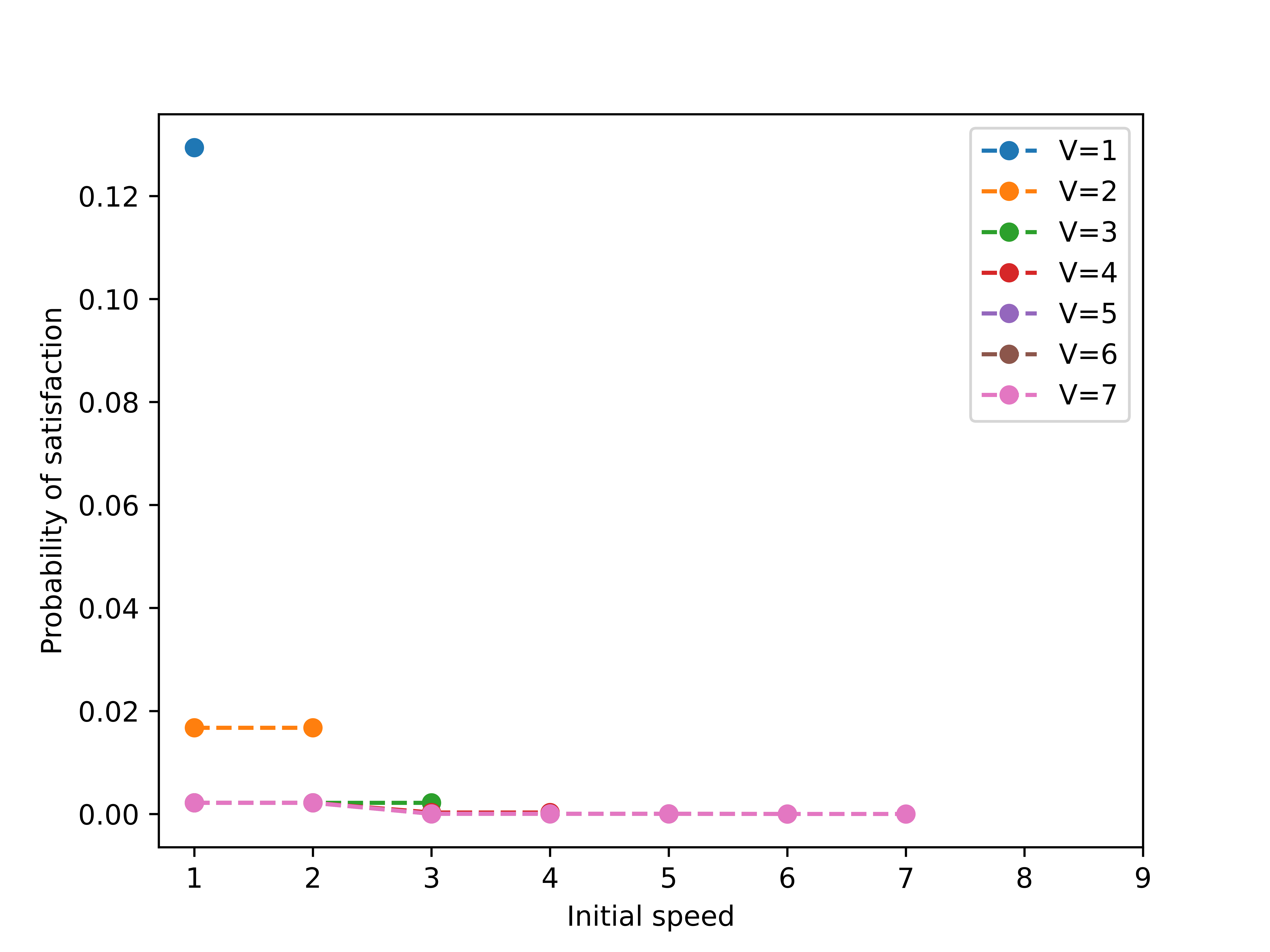}
       \caption{\footnotesize Proposition-labeled confusion matrix}
       \label{fig:prop_plot}
   \end{subfigure} 
\hfill
\begin{subfigure}{0.45\columnwidth}
       \centering
       \includegraphics[width=\linewidth]{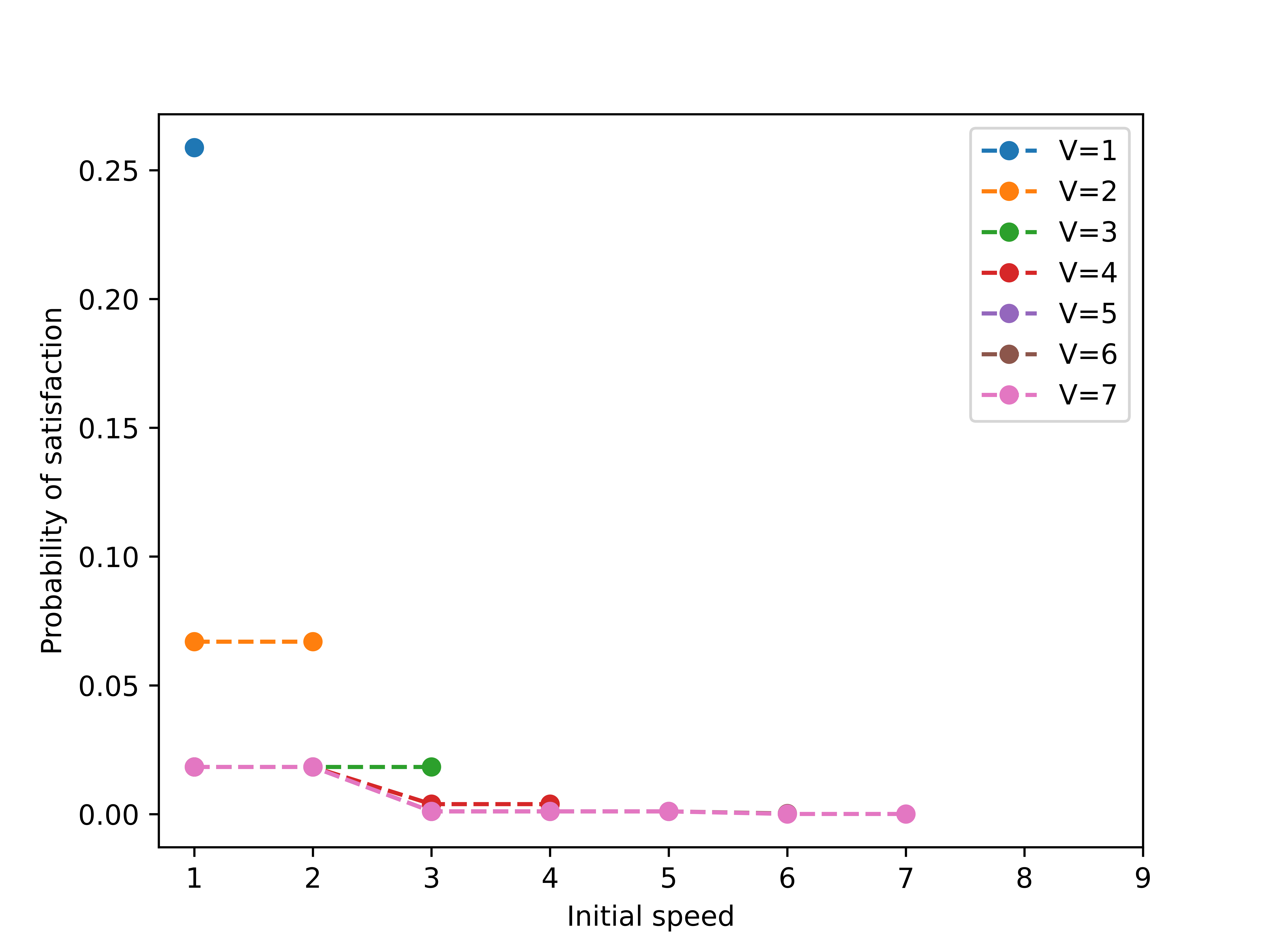}
       \caption{\footnotesize Proposition-labeled confusion matrix parametrized by distance}
       \label{fig:dist_prop_plot}
   \end{subfigure} 
\label{}
\caption{\small Satisfaction probabilities \(\mathbb{P}_{\mM}(s_0 \models \varphi)\) for the car pedestrian running example with Markov chains constructed \(\mM\) constructed from the various confusion matrices. Each plot shows the satisfaction probability as a function of initial speed, with maximum speed as the legend.}
\end{figure}
The satisfaction probabilities of safety requirements are relatively low at around \(20\%\). This is for several reasons. First, we evaluated a model trained with one modality (2D object detection with vision); typically the best models are multi-modal and use data from several different sensors to make predictions. Secondly, we used a pre-trained model that was not trained on the NuScenes dataset. Finally, we do not consider tracking in our evaluation. However, since training object detection models is not the central focus of this work, we chose to evaluate a pre-trained model trained on a different dataset. Yet, we still see insightful differences in the temporal logic analysis of evaluating probability of satisfaction resulting from the choice of confusion matrix. 
Between canonical class based confusion matrix and its distance parametrized counterpart (see Figures~\ref{fig:class_plot} and~\ref{fig:dist_class_plot}), we see a two-fold increase in satisfaction probability \(\mathbb{P}_{\mM}(s_0 \models \varphi)\) for low speeds. Similar trends hold for for proposition-based confusion matrix and its distance parameterized variants as seen in Figures~\ref{fig:prop_plot} and~\ref{fig:dist_prop_plot}. Across all confusion matrices, satisfaction probability decreases with speed, corresponding to not being able to recover from misdetections at higher speed, which is due to our choice of controller.
Lastly, the proposition-labeled confusion matrix results in higher satisfaction probabilities than its class-based counterpart.
\vspace{-1mm}
\section{Conclusion}
\vspace{-1mm}
We introduced two evaluation metrics, a \emph{proposition-labeled} confusion matrix and a \emph{class-labeled distance parameterized} confusion matrix, for object detection tasks. Parameterizing the confusion matrix with distance accounts for differences in detection performance in the temporal logic analysis with the planning module. We empirically observed that the proposition-based confusion matrix resulted in higher satisfaction probability. Additionally, the distance-based metrics give a more optimistic probability of satisfaction. As a next step, we would like to conduct this analysis on richer object detection models trained using LIDAR and RADAR modalities in addition to vision, and also incorporate richer planners designed for lower levels of abstraction of the system. Future work also includes i) studying other perception tasks such as localization and tracking in the context of system-level analysis with respect to formal requirements, ii) extending the analysis in this framework to account for reactive environments, and iii) validating the temporal logic analysis presented in the paper with experiments on hardware.
\clearpage
\bibliographystyle{IEEEtran}
\bibliography{refs}
\end{document}